\pdfoutput=1
\documentclass[11pt]{article}

\usepackage{ACL2023}

\usepackage{times}
\usepackage{latexsym}

\usepackage[T1]{fontenc}
\usepackage[utf8]{inputenc}

\usepackage{microtype}

\usepackage{inconsolata}

\usepackage{graphicx}
\usepackage{subfigure}
\usepackage{amsmath}
\usepackage{multirow}
\usepackage{algorithm, algorithmic}
\usepackage{caption}
\usepackage{pifont}
\usepackage{wrapfig,lipsum,booktabs}
\usepackage{lineno}
\usepackage{colortbl}

\usepackage{amsthm,amsmath,amssymb}

\usepackage{mathrsfs}

\usepackage{bbding}
\usepackage{pifont}

\title{Multi-Grained Knowledge Retrieval for End-to-End Task-Oriented Dialog}

\author{Fanqi Wan\textsuperscript{\rm 1}, Weizhou Shen\textsuperscript{\rm 1}, Ke Yang\textsuperscript{\rm 1},
        Xiaojun Quan\textsuperscript{\rm 1}\thanks{$\;\;$Corresponding authors}, Wei Bi\textsuperscript{\rm 2}\textsuperscript{$*$}
         \\ \textsuperscript{\rm 1}School of Computer Science and Engineering, Sun Yat-sen University, China\\ \textsuperscript{\rm 2}Tencent AI Lab \\
         \{wanfq, shenwzh3, yangk59\}@mail2.sysu.edu.cn, \\ quanxj3@mail.sysu.edu.cn, victoriabi@tencent.com 
}

\begin{document}
\maketitle
\begin{abstract}

Retrieving proper domain knowledge from an external database lies at the heart of end-to-end task-oriented dialog systems to generate informative responses. Most existing systems blend knowledge retrieval with response generation and optimize them with direct supervision from reference responses, leading to suboptimal retrieval performance when the knowledge base becomes large-scale. To address this, we propose to decouple knowledge retrieval from response generation and introduce a multi-grained knowledge retriever (MAKER) that includes an entity selector to search for relevant entities and an attribute selector to filter out irrelevant attributes. To train the retriever, we propose a novel distillation objective that derives supervision signals from the response generator. Experiments conducted on three standard benchmarks with both small and large-scale knowledge bases demonstrate that our retriever performs knowledge retrieval more effectively than existing methods. Our code has been made publicly available.\footnote{https://github.com/18907305772/MAKER}
\end{abstract}

\section{Introduction}

When task-oriented dialog (TOD) systems try to accomplish a task such as restaurant reservations and weather reporting for human users, they generally resort to an external knowledge base (KB) to retrieve relevant entity information for generating an informative system response. Conventional pipeline systems comprise several modules such as dialogue state tracking and dialogue policy learning that require annotations for training, where intermediate predictions such as belief state can be used for the retrieval. By contrast, end-to-end task-oriented dialog (E2E-TOD) systems aim to eliminate the dependence on intermediate annotations and generate the response end-to-end~\citep{glmp}. Apparently, knowledge retrieval is at the core of this task, which is non-trivial as no gold labels are available for training a retriever. Arguably, this problem has limited the performance of existing E2E-TOD systems considering that substantial progress has been made in natural language generation.

\begin{figure}[t]
    \centering
    \includegraphics[width=0.95\linewidth]{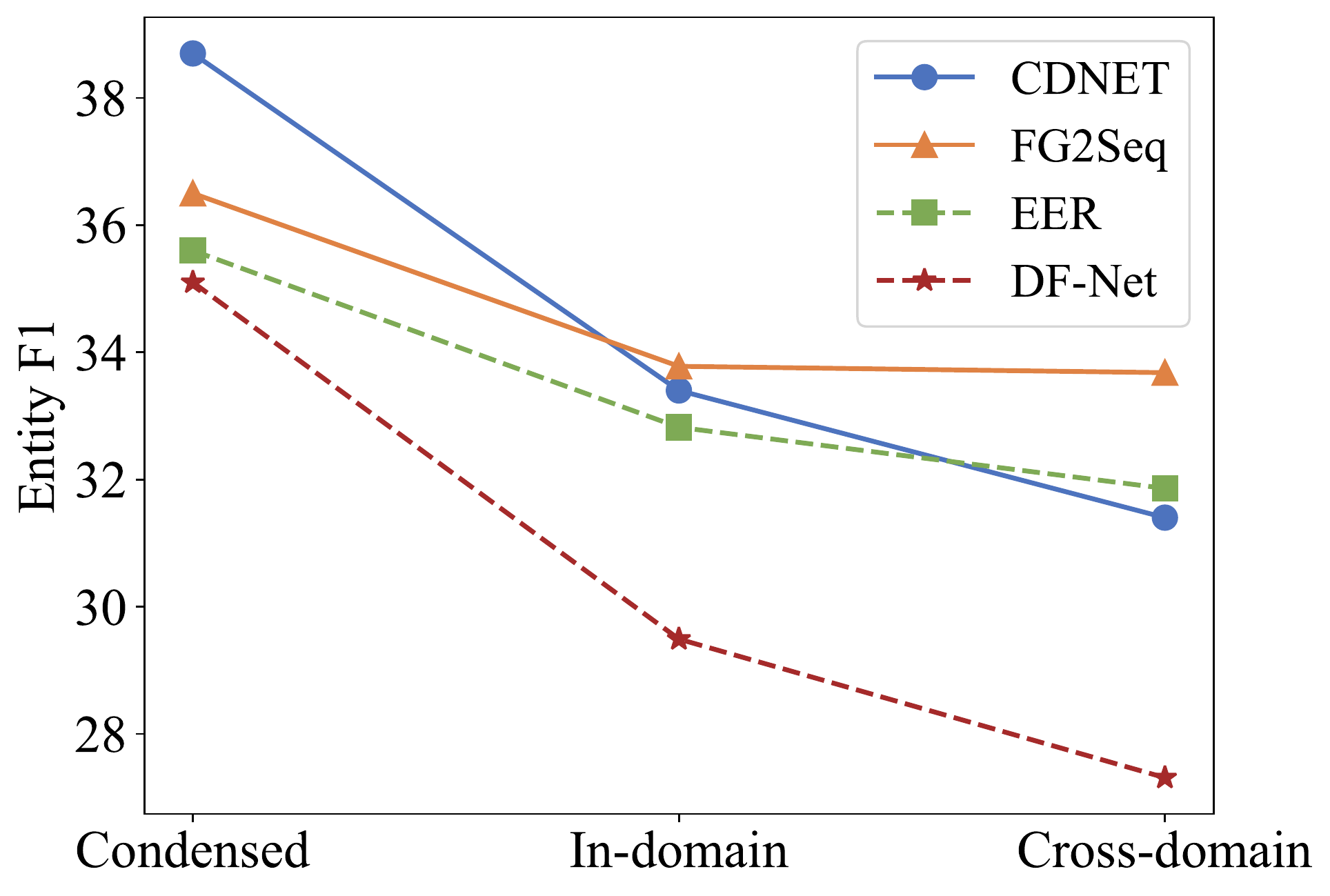}
	\caption{Performance of four end-to-end task-oriented dialog systems on MultiWOZ 2.1 when knowledge bases of different sizes are used. The evaluation metric is Entity F1 scores of entities in generated responses. ``Condensed'' means that each dialog is associated with a small-sized knowledge base, which is the default setting of many current systems. ``In-domain'' means that each dialog corresponds to a knowledge base of the same domain, while ``Cross-domain'' means that all dialogs share the same large-scale cross-domain knowledge base provided in the dataset.}
	\label{fig:ps}
	\vspace{-0.3cm}
\end{figure}

Roughly, existing approaches for knowledge retrieval in E2E-TOD systems can be divided into three categories. First, the knowledge base can be embedded into a memory network and queried with the representations of dialogue context~\citep{mem2seq, dfnet, CDNET}. Second, the serialized knowledge base records can be encoded together with dialog context by pre-trained language models ~\citep{xie2022unifiedskg, wu2022graphmemdialog, q-tod}. Third, the knowledge base can be embedded into model parameters through data augmentation to support implicit knowledge retrieval~\citep{gpt-ke, eco}. These approaches generally blend knowledge retrieval and response generation and train them by the supervision of reference responses, which has two limitations. First, the system response usually consists of pure language tokens and KB-related tokens (e.g., hotel names and phone numbers), and it is challenging to train a good retriever from the weak supervision of reference responses. Second, the systems may become inefficient when the scale of the knowledge base grows large. Our preliminary study\footnote{More details of this study are given in Appendix \ref{sec:ps}.} in Figure \ref{fig:ps} confirms that when a large-scale cross-domain knowledge base is given, existing dialog systems suffer significant performance degradation.

In this paper, we propose a novel Multi-grAined KnowlEdge Retriever (MAKER) for E2E TOD systems to improve the acquisition of knowledge for response generation. The retriever decouples knowledge retrieval from response generation and introduces an entity selector and an attribute selector to select relevant entities and attributes from the knowledge base. Then, the response generator generates a system response based on the dialogue context and the multi-grained retrieval results. The retriever is trained by distilling knowledge from the response generator using the cross-attention scores of KB-related tokens in the response. We train the entity selector, attribute selector, and response generator jointly in an end-to-end manner.

We compare our system with other E2E TOD systems on three benchmark datasets~\citep{smd, camrest, woz}. Empirical results show that our system achieves state-of-the-art performance when either a small or a large-scale knowledge base is used. Through in-depth analysis, we have several findings to report. First, our retriever shows great advantages over baselines when the size of knowledge bases grows large. Second, of the two selectors, the entity selector plays a more important role in the retriever. Third, our system consistently outperforms baselines as different numbers of records are retrieved, and works well even with a small number of retrieval results.

\section{Related Work}

\subsection{End-to-End Task-Oriented Dialog} 

Existing approaches for knowledge retrieval in end-to-end task-oriented dialog systems can be divided into three categories. First, the knowledge base (KB) is encoded with memory networks, and KB records are selected using attention weights between dialogue context and memory cells. Mem2seq~\citep{mem2seq} uses multi-hop attention over memory cells to select KB tokens during response generation. KB-Retriever~\citep{kb-retriever} retrieves the most relevant entity from the KB by means of attention scores to improve entity consistency in the system response. GLMP~\citep{glmp} introduces a global-to-local memory pointer network to retrieve relevant triplets to fill in the sketch response. CDNET~\citep{CDNET} retrieves relevant KB records by computing a distillation distribution based on dialog context.

Second, the concatenation of knowledge base and dialogue context is taken as input for pre-trained language models. UnifiedSKG~\citep{xie2022unifiedskg} uses a unified text-to-text framework to generate system responses. DialoKG~\citep{rony2022dialokg} models the structural information of knowledge base through knowledge graph embedding and performs knowledge attention masking to select relevant triples. Q-TOD~\citep{q-tod} proposes to rewrite dialogue context to  generate a natural language query for knowledge retrieval.

Third, the knowledge base is stored in model parameters for implicit retrieval during response generation. GPT-KE~\citep{gpt-ke} proposes to embed the knowledge base into pre-trained model parameters through data augmentation. ECO~\citep{eco} first generates the most relevant entity with trie constraint to ensure entity consistency in the response. However, these methods generally blend entity retrieval and response generation during response generation, which leads to sub-optimal retrieval performance when large-scale knowledge bases are provided.

\begin{figure*}[t]
    \centering
    \includegraphics[width=0.95\textwidth]{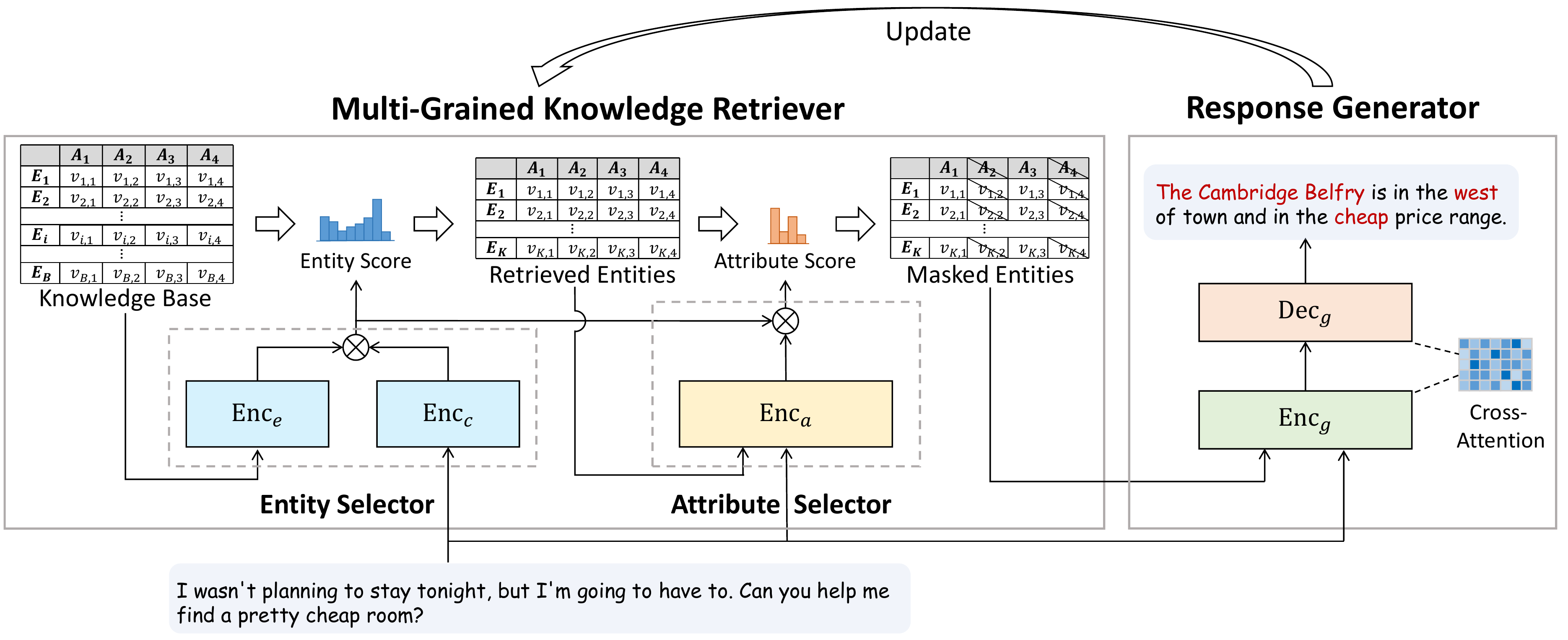}
	\caption{The overview of our end-to-end task-oriented dialog system, which consists of a knowledge retriever and a response generator.  The retriever is further divided into an entity selector and an attribute selector to retrieve multi-grained knowledge, and optimized by distilling knowledge from the response generator.}
	\label{fig:framework}
	\vspace{-0.3cm}
\end{figure*}

\subsection{Neural Retriever} 

With the success of deep neural networks in various NLP tasks, they have also been applied to information retrieval. One of the mainstream approaches is to employ a dual-encoder architecture \citep{dualencoder} to build a retriever. Our work is mostly inspired by the retrieval methods in question answering. To train a retriever with labeled question-document pairs, DPR~\citep{dpr} uses in-batch documents corresponding to other questions together with BM25-retrieved documents as negative samples for contrastive learning. To train a retriever with only question-answer pairs instead of question-document pairs, which is a weakly supervised learning problem, researchers propose to distill knowledge from the answer generator to train the retriever iteratively~\citep{isretriever, fid}. Other researchers try to train the retriever and generator in an end-to-end manner. REALM~\citep{realm}, RAG~\citep{rag}, and EMDR$^2$~\citep{emdr2} propose to train the retriever end-to-end through maximum marginal likelihood. \citet{end2end} propose to combine unsupervised pre-training and supervised fine-tuning to train the retriever. Motivated by these works, we propose a multi-grained knowledge retriever trained by distilling knowledge from the response generator in E2E-TOD systems.

\section{Methods}

In this section, we first describe the notations and outline our method, and then introduce the knowledge retriever and response generator in detail.

\subsection{Notations}

Given a dialog $\mathcal{D}=\{U_{1}, R_{1}, ..., U_{T}, R_{T}\}$ of $T$ turns, where $U_t$ and $R_t$ are the $t$-th turn user utterance and system response, respectively. We use $C_{t}$ to represent the dialog context of the $t$-th turn, where $C_{t}=\{U_{1}, R_{1}, ..., U_{t-1}, R_{t-1}, U_{t}\}$. An external knowledge base (KB) is provided in the form of a set of entities, i.e., $\mathcal{K}=\{E_1, E_2, ..., E_B\}$, where each entity $E_i$ is composed of $N$ attribute-value pairs, i.e., $E_i = \{a^1, v^1_i, ..., a^N, v^N_i\}$. End-to-end task-oriented dialog systems take dialogue context $C_{t}$ and knowledge base $\mathcal{K}$ as input and generate an informative response $R_t$.

\subsection{System Overview}

The architecture of our end-to-end task-oriented dialog system is shown in Figure \ref{fig:framework}.  At each turn of conversation, our system resorts to a Multi-grAined KnowlEdge Retriever (MAKER) to retrieve a set of entities from the external knowledge base.  Then, the response generator takes as input the retrieved entities together with the dialog context and generates a natural language response.~The overall system is optimized in an end-to-end manner without the need for intermediate annotations.

The novelty of MAKER lies in that it decouples knowledge retrieval from response generation and provides multi-grained knowledge retrieval by means of an entity selector and an attribute selector. Specifically, the knowledge base is first encoded with an entity encoder \emph{$\text{Enc}_{e}$} at entity level. Then, the dialogue context is encoded with a context encoder \emph{$\text{Enc}_{c}$} and used to retrieve a set of relevant entities from the knowledge base, which is referred to as entity selection. Next, irrelevant attributes are filtered out with an attribute selector based on the interaction of dialog context and retrieved entities, where another encoder \emph{$\text{Enc}_{a}$} is used. Finally, each retrieved entity is concatenated with the dialog context and passed to a generator encoder \emph{$\text{Enc}_{g}$} to obtain their representations, based on which the generator decoder \emph{$\text{Dec}_{g}$} produces a system response. To train the retriever, the cross-attention scores from KB-related tokens in the reference response to each retrieved entity are used as supervision signals to update the entity selector, while the attribute selector is trained by using the occurrences of attribute values in the dialogue as pseudo-labels. 

To better measure the relationship between entities and response, the whole training process involves two stages. First, the warming-up stage only trains the attribute selector and the response generator, with the entity selector not updated. As the above training converges, the second stage starts to update the entity selector together with other modules using cross-attention scores from the response generator.

\subsection{Knowledge Retriever}
\label{sec: knowledge retriever}

In this section, we introduce the entity selector, attribute selector, and the training of the retriever.

\textbf{Entity Selector} To support large-scale knowledge retrieval, we model the entity selector as a dual-encoder architecture, where one encoder \emph{$\text{Enc}_{c}$} is used to encode the dialogue context and another encoder \emph{$\text{Enc}_{e}$} is to encode each entity (row) of the knowledge base, both into a dense vector. To encode an entity, we concatenate the attribute-value pairs of this entity into a sequence and pass it to \emph{$\text{Enc}_{e}$}. The selection score $s_{t,i}$ for entity $E_{i}$ is defined as the dot product between the context vector and the entity vector as:
\begin{equation}\label{eq1}
s_{t,i} = \text{Enc}_{c}(C_{t})^{T}\text{Enc}_{e}(E_i).
\end{equation}
Then, the top-$K$ entities are obtained by:
\begin{equation}\label{eq2}
\mathcal{E}_{t} = \text{Top}K(s_{t,i}) = \{{E}_1,..., {E}_K\}.
\end{equation}

Retrieving the top-$K$ entities can be formulated as maximum inner product search (MIPS), which can be accelerated to sub-linear time using efficient similarity search libraries such as FAISS ~\citep{faiss}. We implement \emph{$\text{Enc}_{c}$} and \emph{$\text{Enc}_{e}$} with a pre-trained language model and allow them to share weights, where the final ``[CLS]'' token representation is used as the encoder output. Existing studies suggest that initializing \emph{$\text{Enc}_{c}$} and \emph{$\text{Enc}_{e}$} with BERT weights may lead to collapsed representations and harm the retrieval performance. Therefore, following KB-retriever~\citep{kb-retriever}, we initialize them by pre-training with distant supervision.\footnote{More pre-training details are given in Appendix \ref{sec:erp}.}

Since the entity selector is updated by knowledge distillation, recalculating the embeddings of all entities after each update introduces considerable computational cost. Therefore, we follow EMDR$^2$~\citep{emdr2} to update the embeddings of all entities after every 100 training steps.

\textbf{Attribute Selector} To remove irrelevant attributes and values from the retrieved entities for finer-grained knowledge, we design an attribute selector as follows. We first concatenate dialog context $C_{t}$ with each entity ${E}_i \in \mathcal{E}_{t}$ and encode them with an attribute encoder \emph{$\text{Enc}_{a}$}, which is also a pre-trained language model. Then, the final ``[CLS]'' token representation of \emph{$\text{Enc}_{a}$} is extracted and mapped into a $N$-dimensional vector by a feed-forward network (FFN) for attribute scoring:
\begin{equation}\label{eq3}
\mathbf{a}_{t,i} = \text{FFN}(\text{Enc}_{a}([C_{t};{E}_i])),
\end{equation}
where each element in $\mathbf{a}_{t,i} \in \mathbb{R}^N$ represents the importance of the corresponding attribute.

Note that $\mathbf{a}_{t,i}$ only measures the importance of attributes in $E_i$. To obtain the accumulated importance, we calculate the sum of $\mathbf{a}_{t,i}$ over all retrieved entities weighted by entity selection score $s_{t,i}$:
\begin{equation}\label{eq5}
\mathbf{a}_{t} = \sigma(\sum_{i=1}^{K} s_{t,i} \mathbf{a}_{t,i}),
\end{equation}
where $\sigma$ represents the sigmoid function. 

Finally, the attributes whose importance scores in $\mathbf{a}_{t}$ are greater than a pre-defined threshold $\tau$ are selected to construct an attribute subset. The retrieved entities clipped with these attributes are treated as multi-grained retrieval results denoted by $\hat{\mathcal{E}}_{t}$. Specifically, we obtain $\hat{\mathcal{E}}_{t}$ by masking irrelevant attribute-value pairs in each retrieved entity of $\mathcal{E}_t$.
\begin{equation}\label{eq6}
\hat{\mathcal{E}}_{t} = \text{Clip}(\mathcal{E}_{t}, \mathbf{a}_{t}, \tau) = \{\hat{E}_1,..., \hat{E}_K\}.
\end{equation}

To train the attribute selector, we design an auxiliary multi-label classification task. The pseudo-label is a $N$-dimensional 0-1 vector $\mathbf{b}_{t}$ constructed by checking whether any value of an attribute in $\hat{\mathcal{E}}_{t}$ appears in dialogue context $C_t$ or system response $R_t$. Then, we define a binary cross-entropy loss $\mathcal{L}_{att}$ for this classification task as:
\begin{equation}\label{eq7}
\mathcal{L}_{att} = \text{BCELoss}(\mathbf{a}_{t}, \mathbf{b}_{t}).
\end{equation}

\textbf{Updating} The entity selector is updated by distilling knowledge from the response generator as supervision signals. Specifically, since only KB-related tokens in the response are directly connected to the knowledge base, we regard the cross-attention scores from these tokens to each retrieved entity as the knowledge to distill. The rationality behind this is that the cross-attention scores can usually measure the relevance between each entity and the response. Supposing response $R_{t}$ contains $M$ KB-related tokens, we denote the cross-attention scores from each KB-related token to entity $\hat{E}_i$ by $\mathbf{C}_{t,i} \in \mathbb{R}^{|\hat{E}_i|\times M \times L}$, where $|\hat{E}_i|$ represents the number of tokens in $\hat{E}_i$ and $L$ is the number of decoder layers. Then, we calculate an accumulated score for entity $\hat{E}_i$ as:
\begin{equation}
{\hat{c}_{t,i}} = \sum_{j=1}^{|\hat{E}_i|}\sum_{m=1}^{M}\sum_{l=1}^{L}  \mathbf{C}_{t,i,j,m,l}. 
\end{equation}
Then, ${\hat{c}_{t,i}}$ is softmax-normalized to obtain a cross-attention distribution $\mathbf{c}_t$ over the $K$ retrieved entities to reflect their importance for the response. 

Finally, we calculate the KL-divergence between the selection scores $\mathbf{s}_{t}$ of retrieved entities and cross-attention distribution $\mathbf{{c}_{t}}$ as the training loss: 
\begin{equation}\label{eq8}
\mathcal{L}_{ent} = \mathcal{D}_{KL}(\mathbf{s}_{t} || \mathbf{{c}_{t}}).
\end{equation}

\subsection{Response Generator}

Inspired by Fusion-in-Decoder~\citep{fid} in open-domain question answering, we employ a modified sequence-to-sequence structure for the response generator to facilitate direct interaction between dialog context and retrieved entities.

\textbf{Generator Encoder} Each entity $\hat{E}_i$ in $\hat{\mathcal{E}}_{t}$ is first concatenated with dialog context $C_{t}$ and encoded into a sequence of vector representations $\mathbf{H}_{t,i}$:
\begin{equation}\label{eq9}
\mathbf{H}_{t,i} = \text{Enc}_{g}([C_{t};\hat{E_i}]),
\end{equation}
where \emph{$\text{Enc}_{g}$} represents the encoder of the response generator. Then, the representations of all retrieved entities are concatenated into $\mathbf{H}_{t}$:
\begin{equation}\label{eq10}
\mathbf{H}_{t} = [\mathbf{H}_{t,1};...;\mathbf{H}_{t,K}].
\end{equation}

\textbf{Generator Decoder} Taking  $\mathbf{H}_{t}$ as input, the generator decoder \emph{$\text{Dec}_{g}$} produces the system response token by token. During this process, the decoder not only attends to the previously generated tokens through self-attention but also attends to the dialogue context and retrieved entities by cross-attention, which facilitates the generation of an informative response. The probability distribution for each response token in $R_{t}$ is defined as:
\begin{equation}\label{eq11}
P(R_{t,i}) = \text{Dec}_{g}(R_{t,i}|R_{t,<i}, \mathbf{H}_{t}).
\end{equation}

We train the response generator by the standard cross-entropy loss as:
\begin{equation}\label{eq12}
\mathcal{L}_{gen} = \sum_{i=1}^{|R_t|} - \text{log} P(R_{t,i}),
\end{equation}
where $|R_t|$ denotes the length of $R_t$.

Lastly, the overall loss of the system is the sum of entity selection loss $\mathcal{L}_{ent}$, attribute selection loss $\mathcal{L}_{att}$, and response generation loss $\mathcal{L}_{gen}$:
\begin{equation}\label{eq13}
\mathcal{L} = \mathcal{L}_{ent} + \mathcal{L}_{att} + \mathcal{L}_{gen}.
\end{equation}

\subsection{Discussions}
Although deriving much inspiration from open-domain question answering (QA)~\citep{fid}, where the labels for retrieval are also not available, the scenario of this work is quite different. One major difference is that the answer in open-domain QA is completely from the external source of knowledge, while some responses and tokens in dialog systems may not be relevant to the external knowledge base. That means dialog systems need to accommodate both dialog context and external knowledge and generate a fluent and informative natural language response, making this task thornier than open-domain QA.

The main differences between our MAKER and existing knowledge retrieval methods in task-oriented dialog systems are twofold. First, MAKER decouples knowledge retrieval from response generation and provides multi-grained knowledge retrieval of both entities and attributes. The retrieval results are explicitly passed to the generator to produce a system response.~Second, MAKER is trained by distilling knowledge from the response generator for supervision, which varies from existing attention-based approaches.

\section{Experimental Settings}

\subsection{Datasets}
We evaluate our system on three multi-turn task-oriented dialogue datasets: MultiWOZ 2.1 (MWOZ) ~\citep{woz}, Stanford Multi-Domain (SMD) ~\citep{smd}, and CamRest ~\citep{camrest}. Each dialog in these datasets is associated with a condensed knowledge base, which contains all the entities that meet the user goal of this dialog. For MWOZ, each condensed knowledge base contains 7 entities. For SMD and CamRest, the size of condensed knowledge bases is not fixed: it ranges from 0 to 8 with a mean of 5.95 for SMD and from 0 to 57 with a mean of 1.93 for CamRest. We follow the same partitions as previous work~\citep{CDNET}. The statistics of these datasets are shown in Appendix~\ref{sec:statistic}.

BLEU~\citep{bleu} and Entity F1 ~\citep{smd} are used as the evaluation metrics. BLEU measures the fluency of a generated response based on its n-gram overlaps with the gold response. Entity F1 measures whether the generated response contains correct knowledge by micro-averaging the precision and recall scores of attribute values in the generated response.

\subsection{Implementation Details}

 We employ BERT~\citep{bert} as the encoder of our entity selector and attribute selector, and employ T5~\citep{t5} to implement the response generator. All these models are fine-tuned using AdamW optimizer~\citep{adamw} with a batch size of 64. We train these models for 15k gradient steps with a linear decay learning rate of $10^{-4}$. We conduct all experiments on a single 24G NVIDIA RTX 3090 GPU and select the best checkpoint based on model performance on the validation set. More detailed settings can be found in Appendix~\ref{sec:hp}.

\subsection{Baselines}

We compare our system with the following baselines, which are organized into three categories according to how they model knowledge retrieval.

\textbf{Memory network}: These approaches embed the knowledge base into a memory network and query it with
the representation of dialog context, including DSR~\citep{DSR}, KB-Retriever~\citep{kb-retriever}, GLMP~\citep{glmp}, DF-Net~\citep{dfnet}, EER~\citep{eer}, FG2Seq~\citep{he2020fg2seq}, CDNET~\citep{CDNET}, and GraphMemDialog~\citep{wu2022graphmemdialog}.

\textbf{Direct fusion}: These approaches encode serialized knowledge base records together with dialog context by pre-trained language models, including DialoKG~\citep{rony2022dialokg}, UnifiedSKG~\citep{xie2022unifiedskg}, and Q-TOD~\citep{q-tod}.

\textbf{Implicit retrieval}: These approaches embed the knowledge base into model parameters by data augmentation to provide implicit retrieval during response generation, including GPT-2+KE~\citep{gpt-ke} and ECO~\citep{eco}.

\begin{table*}[t]
	\centering
	\resizebox{0.95\textwidth}{!}{
	\begin{tabular}{lcccccc}
		\toprule
		\multirow{2}*{\textbf{Model}} & \multicolumn{2}{c}{\textbf{MWOZ}}  &\multicolumn{2}{c}{\textbf{SMD}} & \multicolumn{2}{c}{\textbf{CamRest}}\\ 
		%\cline{2-9}
		&\textbf{BLEU}& \textbf{Entity F1}  &\textbf{BLEU}& \textbf{Entity F1} &\textbf{BLEU}& \textbf{Entity F1}\\ \hline \hline
            % \multicolumn{7}{c}{\emph{condensed Knowledge Base}} \\ \hline
		DSR~\citep{DSR}                    & 9.10$^{\ddag}$   & 30.00$^{\ddag}$  & 12.70$^{\dag}$  & 51.90$^{\dag}$  & 18.30$^{\dag}$  & 53.60$^{\dag}$      \\
            KB-Retriever~\citep{kb-retriever}  & -      & -       & 13.90  & 53.70  & 18.50  & 58.60      \\
            GLMP~\citep{glmp}                  & 6.90$^{\ddag}$   & 32.40$^{\ddag}$  & 13.90$^{\ddag}$  & 60.70$^{\ddag}$  & 15.10$\S$  & 58.90$\S$    \\
            DF-Net~\citep{dfnet}               & 9.40   & 35.10  & 14.40  & 62.70  & -      & -        \\
            GPT-2+KE~\citep{gpt-ke}            & 15.05  & 39.58   & 17.35  & 59.78  & 18.00  & 54.85   \\
            EER~\citep{eer}                    & 13.60$^{\S}$  & 35.60$^{\S}$  & 17.20$^{\S}$  & 59.00$^{\S}$  & 19.20$^{\S}$  & 65.70$^{\S}$  \\
            FG2Seq~\citep{he2020fg2seq}        & 14.60$^{\S}$  & 36.50$^{\S}$    & 16.80$^{\S}$  & 61.10$^{\S}$  & 20.20$^{\S}$  & 66.40$^{\S}$    \\
            CDNET~\citep{CDNET}                & 11.90  & 38.70  & 17.80  & 62.90  & 21.80  & 68.60    \\
            GraphMemDialog~\citep{wu2022graphmemdialog}        & 14.90  & 40.20  & 18.80  & 64.50  & 22.30  & 64.40    \\
            ECO~\citep{eco}                    & 12.61  & 40.87  & -  & -  & 18.42  & 71.56    \\
            DialoKG~\citep{rony2022dialokg}    & 12.60  & 43.50  & 20.00  & 65.90  & 23.40  & \textbf{75.60}    \\
            UnifiedSKG (T5-Base)~\citep{xie2022unifiedskg}  & -      & -  & 17.41  & 66.45  & -      & -          \\
            UnifiedSKG (T5-Large)~\citep{xie2022unifiedskg}  & 13.69$^{*}$      & 46.04$^{*}$  & 17.27  & 65.85  & 20.31$^{*}$      & 71.03$^{*}$         \\
            Q-TOD (T5-Base)~\citep{q-tod}       & -      & -  & 20.14  & 68.22  & -      & -          \\
            Q-TOD (T5-Large)~\citep{q-tod}      & \underline{17.62}  & 50.61  & 21.33  & \underline{71.11}  & 23.75  & 74.22   \\ \hline
            Ours (T5-Base)                       & 17.23 & \underline{53.68}  & \underline{24.79} & 69.79 & \underline{25.04} & 73.09   \\
            Ours (T5-Large)                      & \textbf{18.77} & \textbf{54.72}  & \textbf{25.91} & \textbf{71.30} & \textbf{25.53} & \underline{74.36}   \\
		\bottomrule
	\end{tabular}
	}
        \caption{Overall results of E2E TOD systems with condensed knowledge bases on MWOZ, SMD, and CamRest. The best scores are highlighted in bold, and the second-best scores are underlined. $\dag$, $\ddag$, $\S$, $*$ indicates that the results are cited from \citep{kb-retriever}, \citep{dfnet}, \citep{CDNET}, and \citep{q-tod}, respectively.}
	\label{tab:main_res_condensed_kb}
	%\vspace{-0.3cm}
\end{table*}

\section{Results and Analysis}
In this section, we first show the overall performance of the evaluated systems given a condensed knowledge base for each dialog. Then, we compare them with a more practical setting in which a large-scale knowledge base is provided. We also conduct an in-depth analysis of the proposed retriever. More experiments are presented in the appendix.

\subsection{Overall Results}\label{sec:overall_res}

The overall results are shown in Table~\ref{tab:main_res_condensed_kb}. We observe that our system with T5-Large as the backbone model achieves the state-of-the-art (SOTA) performance on MWOZ and SMD. Specifically, on MWOZ our system surpasses the previous SOTA, namely Q-TOD, by 1.15 points in BLEU and 4.11 points in Enity F1. On SMD, the improvements over Q-TOD are 4.58 points in BLEU and 0.19 points in Enity F1. On CamRest, our system only achieves the best performance in BLEU but underperforms the best-performing DialoKG slightly. The reason behind this phenomenon is that many dialogues in CamRest contain extremely small knowledge bases, with only 1-2 entities, leaving little space for our retriever to show its advantage.

Note that with the same backbone generator (T5-Base/T5-Large), our system surpasses Q-TOD even though it relies on human annotations to train a query generator for knowledge retrieval. The possible reason is that while the retriever of Q-TOD is independent of response generation, ours is trained and guided by knowledge distillation from response generation. Moreover, in addition to retrieving entities from the knowledge base, our retriever also conducts a fine-grained attribute selection.

\begin{table}[t]
 \centering
    %\vspace{-0.3cm}
	\resizebox{0.999\linewidth}{!}{
	\begin{tabular}{lcccc}
		\toprule
		\multirow{2}*{\textbf{Model}}  & \multicolumn{2}{c}{\textbf{MWOZ}} &\multicolumn{2}{c}{\textbf{CamRest}} \\ 
		%\cline{2-9}
		&\textbf{BLEU}& \textbf{Entity F1} &\textbf{BLEU}& \textbf{Entity F1}\\ \hline \hline
            % \multicolumn{5}{c}{\emph{Full Knowledge Base}} \\ \hline
            DF-Net   & 6.45 & 27.31 & - & -  \\
            EER    & 11.60 & 31.86 & 20.61 & 57.59  \\
            FG2Seq  & 10.74 & 33.68 & 19.20 & 59.35  \\
            CDNET   & 10.90 & 31.40 & 16.50 & 63.60  \\
            Q-TOD & \underline{16.67} & 47.13 & 21.44 & 63.88  \\ \hline
            Ours (T5-Base) & 16.25 & \underline{50.87} & \textbf{26.19} & \underline{72.09}  \\
            Ours (T5-Large) & \textbf{18.23} & \textbf{52.12} & \underline{25.34} & \textbf{72.43} \\
            		\bottomrule
	\end{tabular}
	}
        \caption{Overall results of E2E TOD systems with a large-scale knowledge base on MWOZ and CamRest, respectively. The best scores are highlighted in bold, and the second-best scores are underlined.}
	\label{tab:main_res_full_kb}
	%\vspace{-0.3cm}
\end{table}

\subsection{Large-Scale Knowledge Base}

The experiments in Section~\ref{sec:overall_res} are conducted with each dialog corresponding to a condensed knowledge base. Although most previous systems are evaluated in this setting, it is not practical to have such knowledge bases in real scenes, where the systems may need to retrieve knowledge from a large-scale knowledge base. Therefore, we examine the performance of several well-recognized E2E TOD systems by implementing them on a large-scale cross-domain knowledge base (referred to as ``full knowledge base'') on MWOZ and CamRest, respectively, where the knowledge base is constructed by gathering the entities for all dialogs in the original dataset.\footnote{Since the training scripts of Q-TOD is not released, we directly use its open-source checkpoint (T5-Large) and conduct inference with the full knowledge base.} The results are shown in Table~\ref{tab:main_res_full_kb}.

We observe that our system outperforms baselines by a large margin when the full knowledge base is used. In addition, there are two other observations. First, comparing the results in Table~\ref{tab:main_res_condensed_kb} and Table~\ref{tab:main_res_full_kb}, we note existing systems suffer a severe performance deterioration when the full knowledge base is used. For example, the Enity F1 score of DF-Net drops by 7.79 points on MWOZ, while our system only drops by 2.81/2.6 points. Second, our system with the full knowledge base still outperforms other systems when they use condensed knowledge bases, which is easier to retrieve. These observations verify the superiority of our system when applied to a large-scale knowledge base as well as the feasibility of applying it to real scenes.

\begin{table}[t]
 \centering
    %\vspace{-0.3cm}
	\resizebox{0.98\linewidth}{!}{
	\begin{tabular}{lll}
		\toprule
		\textbf{Model} &\textbf{BLEU}& \textbf{Entity F1} \\ \hline \hline
		%\cline{2-9}
            Ours$_\text{condensed}$   & 17.23 & 53.68 \\
            % \ \ w/o attribute selector & 16.17(\small{$\downarrow$5.43\%})& 51.45(\small{$\downarrow$ 2.61\%}) \\ \hline
            \ \ \emph{w/o}  distillation & 16.21 (\small{$\downarrow$1.02})& 51.05 (\small{$\downarrow$2.63}) \\
            \ \ \emph{w/o} attr\_selector & 15.72 (\small{$\downarrow$1.51})& 51.76 (\small{$\downarrow$1.92}) \\
            \ \ \emph{w/o} ent\_selector & 16.07 (\small{$\downarrow$1.16})& 50.67 (\small{$\downarrow$3.01}) \\ \hline
            Ours$_\text{full}$   & 16.25 & 50.87 \\
            \ \ \emph{w/o} distillation & 15.85 (\small{$\downarrow$0.40}) & 48.28 (\small{$\downarrow$2.59}) \\
            \ \ \emph{w/o} attr\_selector & 15.40 (\small{$\downarrow$0.85}) & 48.55 (\small{$\downarrow$2.32}) \\
		\bottomrule
	\end{tabular}
	}
 \caption{Results of ablation study on MWOZ with T5-base, where ``w/o'' means without, ``distillation'' denotes distillation from response generation, ``attr\_selector'' denotes the attribute selector, and ``ent\_selector'' denotes the entity selector.}
	\label{tab:ablation_res}
	%\vspace{-0.3cm}
\end{table}

\subsection{Ablation Study}

We conduct an ablation study of our retriever MAKER with both condensed and full knowledge bases on MWOZ, and show the results in the first and the second blocks of Table \ref{tab:ablation_res}, respectively. 

When condensed knowledge bases are used, the system suffers obvious performance drops with the removal of distillation (\emph{w/o} distillation) or entity selection (\emph{w/o} ent\_selector). This indicates that despite the quality of condensed knowledge bases, our retriever can further learn to distinguish between the entities by distilling knowledge from the response generator. Besides, the performance of the system drops when the attribute selector is abandoned (\emph{w/o} attr\_selector), showing that attribute selection is also indispensable in the retriever.

When the full knowledge base is used, entity selection is more necessary for the system. Therefore, we only ablate the distillation component and the attribute selector. The results show that the system suffers significant performance degradation when distillation is removed (\emph{w/o} distillation). Attribute selection is also shown important as the performance drops upon it is removed (\emph{w/o} attr\_selector).

\begin{table}[t]
 \centering
    %\vspace{-0.3cm}
	\resizebox{0.98\linewidth}{!}{
	\begin{tabular}{lccc}
		\toprule
		\textbf{Retrieval Method} &\textbf{BLEU}& \textbf{Entity F1} & \textbf{Recall@7} \\ \hline \hline
		%\cline{2-9}
            Oracle   & 16.17 & 51.45 & 100.00 \\
            MAKER    & 17.18 & 49.05 & 86.47 \\
            Pre-training     & 16.67 & 48.77 & 82.71 \\
            Frequency    & 16.60 & 48.00 & 75.94 \\
            BM25    & 16.21 & 45.56 & 26.32 \\
		\bottomrule
	\end{tabular}
	}
        \caption{Comparison of different retrieval methods on the full knowledge base. \textit{Oracle} refers to using the condensed knowledge base for each dialog as the retrieval result. \textit{Frequency} means measuring the relevance by the frequency of attribute values occurring in the dialogue context. \textit{BM25} measures the relevance using the BM25 score between dialogue context and each entity.}
	\label{tab:entity_retrieval_res}
	%\vspace{-0.3cm}
\end{table}

\begin{figure}[t] \centering    
\subfigure[Recall] { 
\label{fig:retriever-recall}     
\includegraphics[width=0.46\linewidth]{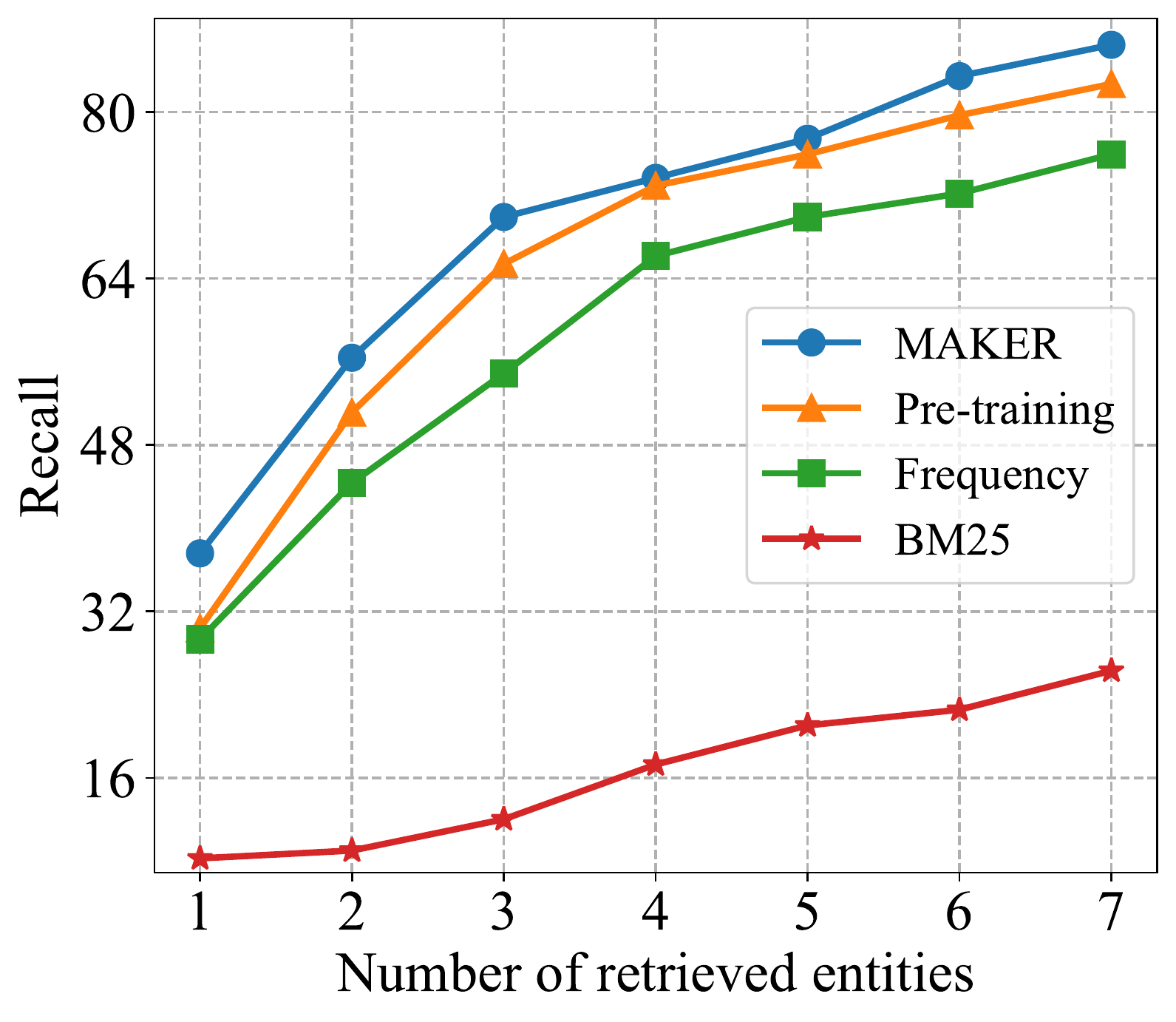}     
}   
\subfigure[Entity F1] {
 \label{fig:retriever-f1}     
\includegraphics[width=0.46\linewidth]{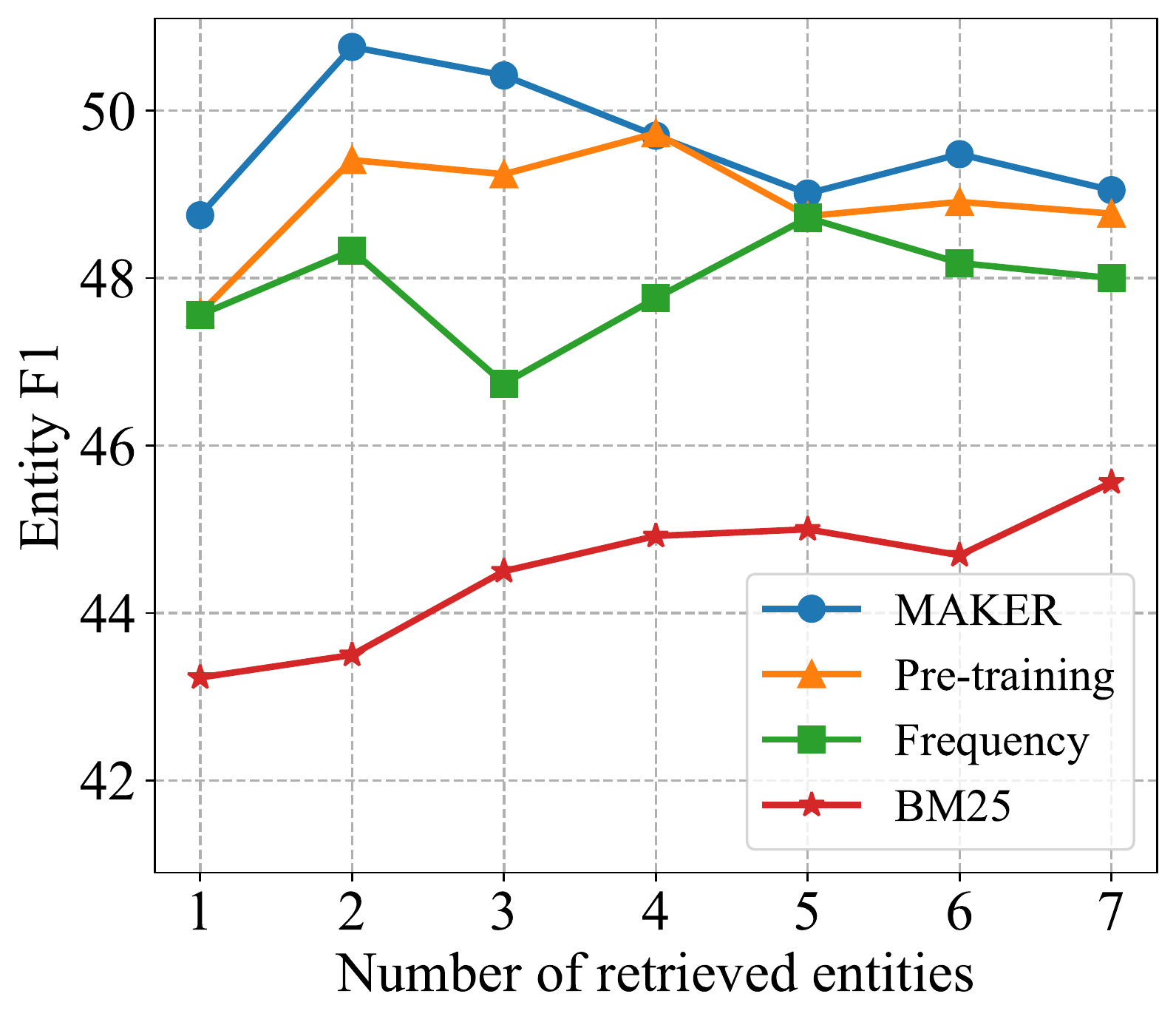}  
}     
\caption{Performance of different retrieval methods as the number of retrieved entities changes on the full knowledge base in Recall (a) and Entity F1 (b) scores.}     
\label{fig:retriever-f1-recall} 
\vspace{-0.2cm}
\end{figure}

\subsection{Comparison of Retrieval Methods}

To further demonstrate the effectiveness of our multi-grained knowledge retriever, we compare different retrieval methods on the full knowledge base of MWOZ. Specifically, we first retrieve the top-$K$ entities with different retrieval methods and employ the same response generator to generate the system response. Moreover, we propose a new metric, i.e., Recall@7, to measure whether the suggested entities in the system response appear in the 7 retrieved entities. As shown in Table \ref{tab:entity_retrieval_res}, the proposed retriever achieves the best performance compared with other methods except Oracle, which uses condensed knowledge bases without retrieval, in both generation metrics (BLEU, Entity F1) and the retrieval metric (Recall@7). 

To investigate the effect of different numbers of retrieved entities on system performance, we report the Entity F1 and Recall@$x$ scores of the above retrieval methods as the number of entities changes, while Oracle is not included because we cannot rank its entities. We observe in Figure \ref{fig:retriever-recall} that the Recall@$x$ scores for all methods improve as the number of entities grows, while our retriever consistently achieves the best performance. In Figure \ref{fig:retriever-f1}, we observe no positive correlation between the Entity F1 score and the number of entities, suggesting that noisy entities may be introduced as the number of entities increases. We can also observe that the number of entities corresponding to the peak of the Entity F1 scores varies for different methods, while our retriever only requires a small number of entities to reach the peak performance.

\subsection{Attribute Selection Methods}

In Section \ref{sec: knowledge retriever}, we calculate an accumulated importance score for each attribute weighted by entity selection scores to determine which attributes are preserved based on a given threshold. In Table \ref{tab:attribute_selection_res}, we compare different methods for accumulating the attribute scores as well as different approaches for filtering out irrelevant attributes. It can be observed that direct averaging rather than weighting by entity selection scores hurts the Entity F1 score. This indicates that the retriever can select attributes more appropriately based on the selection scores of retrieved entities. We also observe that using top-$K$ instead of a threshold to select attributes leads to a lower Entity F1 score than preserving all attributes. We believe the reason is that the number of attributes to be selected varies for each dialogue context, and therefore simply selecting the top-$K$ attributes results in sub-optimal attributes.

\begin{table}[t]
 \centering
    %\vspace{-0.3cm}
	\resizebox{0.6\linewidth}{!}{
	\begin{tabular}{lcc}
		\toprule
		\textbf{Method} &\textbf{BLEU}& \textbf{Entity F1} \\ \hline \hline
		%\cline{2-9}
            Weighted   & 16.25 & 50.87 \\
            Average     & 16.46 & 48.81 \\ \hline
            Threshold     & 16.25 & 50.87 \\
            Top-$K$     & 16.31 & 46.89 \\
            All     & 15.40 & 48.55 \\
		\bottomrule
	\end{tabular}
	}
        \caption{Comparison of attribute selection methods for MAKER on the full knowledge base. The upper two rows are methods for accumulating attribute importance scores across retrieved entities, and the bottom three rows are methods for filtering out irrelevant attributes.}
	\label{tab:attribute_selection_res}
	%\vspace{-0.3cm}
\end{table}

\section{Conclusion}

We propose a novel multi-grained knowledge retriever (MAKER) for end-to-end task-oriented dialog systems.~It decouples knowledge retrieval from response generation and introduces an entity selector and an attribute selector to acquire multi-grained knowledge from the knowledge base. The retriever is trained by distilling knowledge from the response generator. Empirical results show that our system achieves state-of-the-art performance when either a small or a large-scale knowledge base is provided for each dialog. Through in-depth analysis, our retriever shows great advantages over baselines when the size of knowledge bases grows large. Of the two selectors, the entity selector is shown to be more prominent in the retriever. 

\section*{Acknowledgements}
This work was supported by the National Natural Science Foundation of China (No.62176270), the Guangdong Basic and Applied Basic Research Foundation (No.2023A1515012832), the Program for Guangdong Introducing Innovative and Entrepreneurial Teams (No. 2017ZT07X355), and the Tencent AI Lab Rhino-Bird Focused Research Program. We thank Yingqi Gao and Canbin Huang for their efforts in the preliminary experiments.

\section*{Limitations}

Our system employs a modified sequence-to-sequence architecture to implement the response generator. Since the length of dialogue context increases as the dialogue continues, the generator needs to input multiple long dialogue contexts to the encoder simultaneously, each for a retrieved entity. This may cause redundancy in the input and lowers the proportion of KB-related information. We will explore more efficient architectures for the response generator in future work.

\section*{Ethics Statement}
All the experiments are conducted on publicly available datasets, which don't include any private information. Our work doesn't involve identity characteristics or any gender and racial discrimination. 

% \section*{Acknowledgements}
% This document has been adapted by Jordan Boyd-Graber, Naoaki Okazaki, Anna Rogers from the style files used for earlier ACL, EMNLP and NAACL proceedings, including those for
% EACL 2023 by Isabelle Augenstein and Andreas Vlachos,
% EMNLP 2022 by Yue Zhang, Ryan Cotterell and Lea Frermann,
% ACL 2020 by Steven Bethard, Ryan Cotterell and Rui Yan,
% ACL 2019 by Douwe Kiela and Ivan Vuli\'{c},
% NAACL 2019 by Stephanie Lukin and Alla Roskovskaya, 
% ACL 2018 by Shay Cohen, Kevin Gimpel, and Wei Lu, 
% NAACL 2018 by Margaret Mitchell and Stephanie Lukin,
% Bib\TeX{} suggestions for (NA)ACL 2017/2018 from Jason Eisner,
% ACL 2017 by Dan Gildea and Min-Yen Kan, NAACL 2017 by Margaret Mitchell, 
% ACL 2012 by Maggie Li and Michael White, 
% ACL 2010 by Jing-Shin Chang an d Philipp Koehn, 
% ACL 2008 by Johanna D. Moore, Simone Teufel, James Allan, and Sadaoki Furui, 
% ACL 2005 by Hwee Tou Ng and Kemal Oflazer, 
% ACL 2002 by Eugene Charniak and Dekang Lin, 
% and earlier ACL and EACL formats written by several people, including
% John Chen, Henry S. Thompson and Donald Walker.
% Additional elements were taken from the formatting instructions of the \emph{International Joint Conference on Artificial Intelligence} and the \emph{Conference on Computer Vision and Pattern Recognition}.

\bibliography{anthology,custom}
\bibliographystyle{acl_natbib}

\vspace{0.3cm}
\appendix

\section{Statistics of Datasets}

The statistics of the datasets are shown in Table \ref{tab:dataset statistics}.

\label{sec:statistic}

\begin{table}[h]
	\centering
	\resizebox{0.99\linewidth}{!}{
	\begin{tabular}{lccccc}
		\toprule
  \multirow{2}*{\textbf{Dataset}} &\multirow{2}*{\textbf{Domains}}& \textbf{\# Dialogues} \\ \cline{3-3}
  && \textbf{Train/Val/Test} \\ \hline \hline
		%\cline{2-9}
            MWOZ~\citep{woz}                  & Restaurant, Attraction, Hotel  & 1839/117/141    \\
		SMD~\citep{smd}                   & Navigate, Weather, Schedule  & 2425/302/304    \\
            CamRest~\citep{camrest}           & Restaurant  & 406/135/135    \\
		\bottomrule
	\end{tabular}
	}
        \caption{Statistics of the three datasets.}
	\label{tab:dataset statistics}
	%\vspace{-0.3cm}
\end{table}

\section{Preliminary Study}
\label{sec:ps}

The detailed results of our preliminary study for condensed, in-domain, and cross-domain knowledge bases are shown in Table \ref{tab:preliminary_res}. The results of baseline models on condensed knowledge bases are cited from \citep{CDNET}. We produce their results on in-domain and cross-domain knowledge bases by using the officially released code.

\begin{table}[h]
 \centering
    %\vspace{-0.3cm}
	\resizebox{0.99\linewidth}{!}{
	\begin{tabular}{lcccccc}
		\toprule
		\multirow{2}*{\textbf{Model}}  &\multicolumn{2}{c}{\textbf{Condensed}} & \multicolumn{2}{c}{\textbf{In-Domain}} & \multicolumn{2}{c}{\textbf{Cross-Domain}} \\ 
		%\cline{2-9}
		&\textbf{BLEU}& \textbf{Entity F1} &\textbf{BLEU}& \textbf{Entity F1} &\textbf{BLEU}& \textbf{Entity F1}\\ \hline \hline
		%\cline{2-9}
            DF-Net & 9.40  & 35.10  & 7.24  & 29.49  & 6.45   & 27.31    \\
            EER & 13.60  & 35.60  & 11.44  & 32.82  & 11.60   & 31.86    \\
            FG2Seq & 14.60  & 36.50  & 10.53  & 33.78  & 10.74   & 33.68    \\
            CDNET & 11.90  & 38.70  & 11.70  & 33.40  & 10.90   & 31.40    \\
		\bottomrule
	\end{tabular}
	}
        \caption{Comparison of end-to-end task-oriented dialog systems with different sizes of knowledge bases.}
	\label{tab:preliminary_res}
	%\vspace{-0.3cm}
\end{table}

\section{Pre-training for Entity Selector}
\label{sec:erp}

Given a dialogue context and the system response, we use the entity with the most occurrences of its attribute values in the dialogue context and system response as the label. Then we apply supervised contrastive learning for optimization~\citep{simcse}. Specifically, the positive example of a dialogue context is the corresponding labeled entity, while the negative examples are the labeled entities of other examples in the same mini-batch. Then, we employ the InfoNCE loss as the training objective to pull close the sentence representations of positive samples and push away that of negative samples. We conduct this pre-training on the MWOZ and CamRest datasets. Since the knowledge base of the SMD dataset is strictly specific to each dialog, we cannot collect a global knowledge base from the dialogs. Thus, the pre-training is not conducted on SMD. The hyperparameters for the pre-training are shown in Table \ref{tab:hyper_pretrain}.

\begin{table}[hb]
 \centering
    %\vspace{-0.3cm}
	\resizebox{0.9\linewidth}{!}{
	\begin{tabular}{lcc}
		\toprule
		\textbf{Hyperparameters} &\textbf{MWOZ}& \textbf{CamRest} \\ \hline \hline
		%\cline{2-9}
            Optimizer   & AdamW & AdamW \\
            Batch size & 128 & 108 \\
            Epoch & 10 & 15 \\
            Learning rate schedule & Linear & Linear \\
            Learning rate & 5e-5 & 5e-5 \\
            Weight decay & 0.01 & 0.01 \\
            Temperature & 0.05 & 0.05 \\
            Max length & 128 & 128 \\
            Pooler type & CLS & CLS \\
            Pooler dimension & 128 & 128 \\
		\bottomrule
	\end{tabular}
	}
        \caption{Hyperparameter setting for pre-training our entity selector on the full knowledge base of MWOZ and CamRest datasets, respectively.}
	\label{tab:hyper_pretrain}
	%\vspace{-0.3cm}
\end{table}

\section{Domain-Wise Results}
\label{sec:dwr}

\begin{table*}[t]
 \centering
    %\vspace{-0.3cm}
	\resizebox{0.65\textwidth}{!}{
	\begin{tabular}{lccccc}
		\toprule
		\textbf{Model} &\textbf{BLEU}& \textbf{Entity F1} & \textbf{Hotel} & \textbf{Attraction} & \textbf{Restaurant} \\ \hline \hline
		%\cline{2-9}
            DSR   & 9.10 & 30.00 & 27.10 & 28.00 & 33.40 \\
            GLMP   & 6.90 & 32.40 & 28.10 & 24.40 & 38.40 \\
            DF-Net   & 9.40 & 35.10 & 30.60 & 28.10 & 40.90 \\
            GPT-2+KE   & 15.00 & 39.60 & 33.40 & 43.30 & 37.10 \\
            EER   & 13.60 & 35.60 & 35.70 & 43.00 & 34.20 \\
            FG2Seq   & 14.60 & 36.50 & 34.40 & 37.20 & 38.90 \\
            CDNET   & 11.90 & 38.70 & 36.30 & 38.90 & 41.70 \\
            GraphMemDialog   & 14.90 & 40.20 & 36.40 & 48.80 & 42.80 \\
            DialoKG   & 12.60 & 43.50 & 37.90 & 39.80 & 46.70 \\
            Q-TOD (T5-Large)   & \underline{17.62} & \underline{50.61} & \underline{45.25} & \underline{54.81} & \underline{55.78} \\ \hline
            Ours (T5-Large)   & \textbf{18.77} & \textbf{54.72} & \textbf{46.97} & \textbf{65.08} & \textbf{62.12} \\
		\bottomrule
	\end{tabular}
	}
        \caption{Domain-wise performance on MWOZ.}
	\label{tab:mwoz_domain_res}
	%\vspace{-0.3cm}
\end{table*}

\begin{table*}[t]
 \centering
    %\vspace{-0.3cm}
	\resizebox{0.65\textwidth}{!}{
	\begin{tabular}{lccccc}
		\toprule
		\textbf{Model} &\textbf{BLEU}& \textbf{Entity F1} & \textbf{Schedule} & \textbf{Weather} & \textbf{Navigate} \\ \hline \hline
		%\cline{2-9}
            DSR   & 12.70 & 51.90 & 52.10 & 50.40 & 52.00 \\
            GLMP   & 13.90 & 59.60 & 70.20 & 58.00 & 54.30 \\
            DF-Net   & 14.40 & 62.70 & 73.10 & 57.60 & 57.90 \\
            GPT-2+KE   & 17.40 & 59.80 & 72.60 & 57.70 & 53.50 \\
            EER   & 17.20 & 59.00 & 71.80 & 57.80 & 52.50 \\
            FG2Seq   & 16.80 & 61.10 & 73.30 & 57.40 & 56.10 \\
            CDNET   & 17.80 & 62.90 & 75.40 & 61.30 & 56.70 \\
            GraphMemDialog   & 18.80 & 64.50 & 75.90 & 62.30 & 56.30 \\
            DialoKG   & 20.00 & 65.90 & 77.90 & \textbf{72.70} & 58.40 \\
            Q-TOD (T5-Large)   & \underline{21.33} & \underline{71.11} & \textbf{81.42} & 69.18 & \textbf{62.91} \\ \hline
            Ours (T5-Large)   & \textbf{25.91} & \textbf{71.30} & \underline{78.56} & \underline{72.69} & \underline{62.15} \\
		\bottomrule
	\end{tabular}
	}
        \caption{Domain-wise performance on SMD.}
	\label{tab:smd_domain_res}
	%\vspace{-0.3cm}
\end{table*}

We report the domain-wise results with condensed knowledge bases on MWOZ and SMD in Table \ref{tab:mwoz_domain_res} and Table \ref{tab:smd_domain_res}, respectively. The results of baseline models are cited from \citep{CDNET}, \citep{rony2022dialokg}, and \citep{q-tod}.

\section{More Implementation Details}
\label{sec:hp}

\begin{table*}[t]
 \centering
    %\vspace{-0.3cm}
	\resizebox{0.95\textwidth}{!}{
	\begin{tabular}{lcccccc}
		\toprule
            \multirow{2}*{\textbf{Hyperparameters}}  &\multicolumn{2}{c}{\textbf{MWOZ}} & \multicolumn{2}{c}{\textbf{SMD}} & \multicolumn{2}{c}{\textbf{CamRest}} \\ 
		%\cline{2-9}
		&\textbf{T5-Base} & \textbf{T5-Large} &\textbf{T5-Base}& \textbf{T5-Large} &\textbf{T5-Base}& \textbf{T5-Large} \\ \hline \hline
            Optimizer   & AdamW & AdamW & AdamW & AdamW & AdamW & AdamW \\
            Batch size & 2 & 1 & 2 & 2 & 2 & 2 \\
            Gradient accumulation steps & 32 & 64 & 32 & 32 & 32 & 32  \\
            Training gradient steps & 1500 & 1500 & 1500 & 1500 & 1000 & 1000 \\
            Learning rate schedule & Linear & Linear & Linear & Linear & Linear & Linear \\
            Entity selector learning rate & 5e-5 & 1e-4 & 1e-4 & 1e-4 & 1e-4 & 1e-4 \\
            Attribute selector learning rate & 5e-5 & 1e-4 & 1e-4 & 1e-4 & 1e-4 & 1e-4 \\
            Response generator learning rate & 1e-4 & 1e-4 & 1e-4 & 1e-4 & 1e-4 & 7e-5 \\
            Weight decay & 0.01 & 0.01 & 0.01 & 0.01 & 0.01 & 0.01 \\
            Gradient clipping & 1.0 & 1.0 & 1.0 & 1.0 & 1.0 & 1.0 \\
            Entity selector max length & 128 & 128 & 128 & 128 & 128 & 128 \\
            Attribute selector max context length & 200 & 200 & 200 & 200 & 200 & 200 \\
            Attribute selector max kb length & 100 & 100 & 200 & 200 & 100 & 100 \\
            Response generator max context length & 200 & 200 & 200 & 200 & 200 & 200 \\
            Response generator max kb length & 100 & 100 & 200 & 200 & 100 & 100 \\
            Max output length & 64 & 64 & 128 & 128 & 64 & 64 \\
            Top-$K$ retrieval entities & 6 & 7 & 8 & 8 & 6 & 4 \\
            Attribute selection threshold & 0.1 & 0.1 & 0.0 & 0.0 & 0.0 & 0.0 \\
            Distillation start gradient steps & 625 & 938 & 1500 & 1500 & 750 & 750 \\
		\bottomrule
	\end{tabular}
	}
        \caption{Hyperparameter settings of our system when condensed knowledge bases are used on the MWOZ, SMD, and CamRest datasets.}
	\label{tab:hyper_main_condensed_kb}
	%\vspace{-0.3cm}
\end{table*}

\begin{table*}[thb]
 \centering
    %\vspace{-0.3cm}
	\resizebox{0.80\textwidth}{!}{
	\begin{tabular}{lcccc}
		\toprule
            \multirow{2}*{\textbf{Hyperparameters}}  &\multicolumn{2}{c}{\textbf{MWOZ}} & \multicolumn{2}{c}{\textbf{CamRest}} \\ 
		%\cline{2-9}
		&\textbf{T5-Base} & \textbf{T5-Large} &\textbf{T5-Base}& \textbf{T5-Large} \\ \hline \hline
            Optimizer   & AdamW & AdamW & AdamW & AdamW \\
            Batch size & 2 & 1 & 2 & 1 \\
            Gradient accumulation steps & 32 & 64 & 32 & 64  \\
            Training gradient steps & 1500 & 1500 & 1500 & 1500 \\
            Learning rate schedule & Linear & Linear & Linear & Linear \\
            Entity selector learning rate & 5e-5 & 1e-4 & 1e-4 & 1e-4 \\
            Attribute selector learning rate & 5e-5 & 1e-4 & 1e-4 & 1e-4 \\
            Response generator learning rate & 1e-4 & 1e-4 & 1e-4 & 1e-4 \\
            Weight decay & 0.01 & 0.01 & 0.01 & 0.01 \\
            Gradient clipping & 1.0 & 1.0 & 1.0 & 1.0 \\
            Entity selector max length & 128 & 128 & 128 & 128 \\
            Attribute selector max context length & 200 & 200 & 200 & 200 \\
            Attribute selector max kb length & 100 & 100 & 100 & 100 \\
            Response generator max context length & 200 & 200 & 200 & 200 \\
            Response generator max kb length & 100 & 100 & 100 & 100 \\
            Max output length & 64 & 64 & 64 & 64 \\
            Top-$K$ retrieval entities & 7 & 7 & 7 & 7 \\
            Attribute selection threshold & 0.2 & 0.2 & 0.1 & 0.1 \\
            Distillation start gradient steps & 938 & 938 & 938 & 938 \\
		\bottomrule
	\end{tabular}
	}
        \caption{Hyperparameter settings of our system when the full knowledge base is used on MWOZ and CamRest.}
	\label{tab:hyper_main_full_kb}
	%\vspace{-0.3cm}
\end{table*}

The hyperparameters of our system with condensed and full knowledge bases are shown in Table \ref{tab:hyper_main_condensed_kb} and Table \ref{tab:hyper_main_full_kb}, respectively. 

Our method has three contributions: knowledge distillation, entity selection, and attribute selection. We list the application of these contributions with condensed and full knowledge base in Table \ref{tab:contribution_condensed_kb} and Table \ref{tab:contribution_full_kb}, respectively.

\begin{table*}[t]
 \centering
    %\vspace{-0.3cm}
	\resizebox{0.8\textwidth}{!}{
	\begin{tabular}{lcccccc}
		\toprule
            \multirow{2}*{\textbf{Contributions}}  &\multicolumn{2}{c}{\textbf{MWOZ}} & \multicolumn{2}{c}{\textbf{SMD}} & \multicolumn{2}{c}{\textbf{CamRest}} \\ 
		%\cline{2-9}
		&\textbf{T5-Base} & \textbf{T5-Large} &\textbf{T5-Base}& \textbf{T5-Large} &\textbf{T5-Base}& \textbf{T5-Large} \\ \hline \hline
            Knowledge distillation   & \ding{51} & \ding{51} & \ding{53} & \ding{53} & \ding{51} & \ding{51} \\
            Entity Selection & \ding{51} & \ding{53} & \ding{53} & \ding{53} & \ding{51} & \ding{51} \\
            Attribute Selection & \ding{51} & \ding{51} & \ding{53} & \ding{53} & \ding{53} & \ding{53}  \\
		\bottomrule
	\end{tabular}
	}
        \caption{Hyperparameter settings of whether to apply each contribution to our system when condensed knowledge bases are used on the MWOZ, SMD, and CamRest datasets.}
	\label{tab:contribution_condensed_kb}
	%\vspace{-0.3cm}
\end{table*}

\begin{table*}[thb]
 \centering
    %\vspace{-0.3cm}
	\resizebox{0.6\textwidth}{!}{
	\begin{tabular}{lcccc}
		\toprule
            \multirow{2}*{\textbf{Contributions}}  &\multicolumn{2}{c}{\textbf{MWOZ}} & \multicolumn{2}{c}{\textbf{CamRest}} \\ 
		%\cline{2-9}
		&\textbf{T5-Base} & \textbf{T5-Large} &\textbf{T5-Base}& \textbf{T5-Large} \\ \hline \hline
            Knowledge distillation   & \ding{51} & \ding{51} & \ding{51} & \ding{51} \\
            Entity Selection & \ding{51} & \ding{51} & \ding{51} & \ding{51} \\
            Attribute Selection & \ding{51} & \ding{51} & \ding{51} & \ding{51} \\
		\bottomrule
	\end{tabular}
	}
        \caption{Hyperparameter settings of whether to apply each contribution to our system when the full knowledge base is used on MWOZ and CamRest.}
	\label{tab:contribution_full_kb}
	%\vspace{-0.3cm}
\end{table*}

\section{Case Study}
\label{sec:case}

\begin{figure*}[thb]
    \centering
    \includegraphics[width=0.99\textwidth]{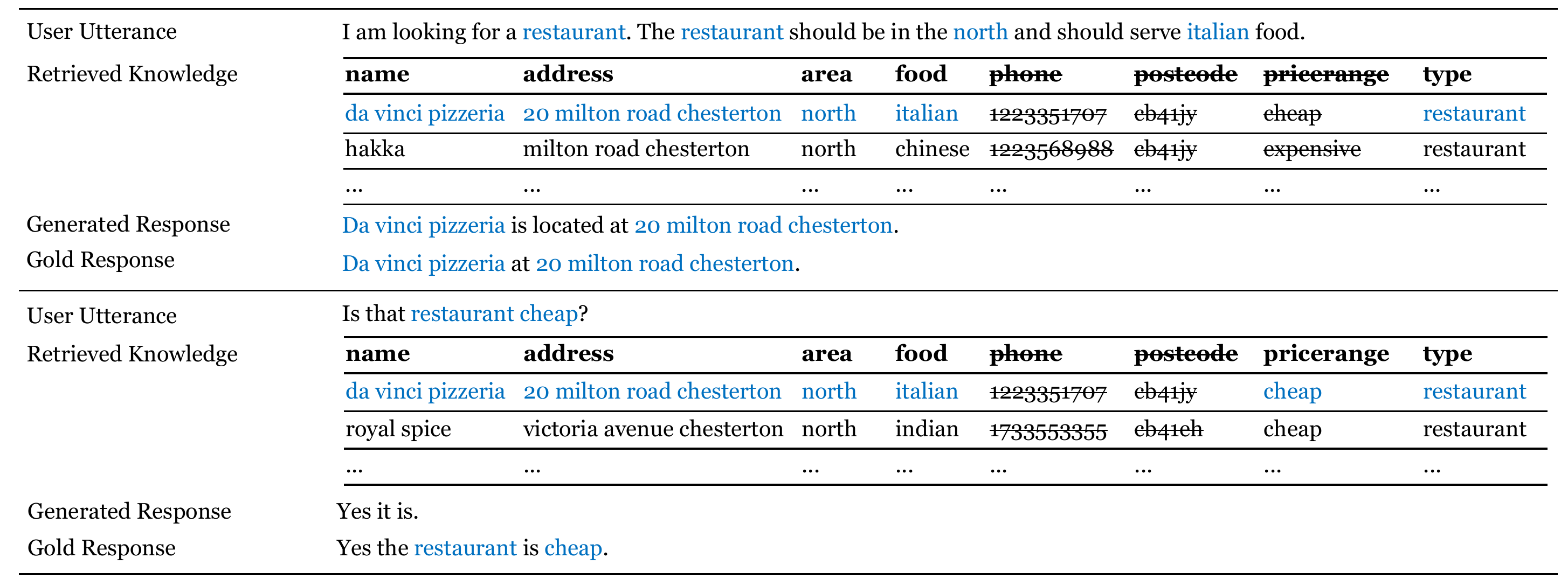}
	\caption{An example of dialogue to illustrate our system. Blue font refers to knowledge base-related information.}
	\label{fig:case_study}
	\vspace{-0.3cm}
\end{figure*}

In Figure \ref{fig:case_study}, we provide a dialogue example from the MWOZ dataset. We can observe that, for a given user utterance, our system can retrieve entities that satisfy the user goal, while masking irrelevant attributes. Then, it generates appropriate system responses. Note that when the user goal changes, e.g., in the second turn of this case when the user wants a cheap restaurant, our retriever can retrieve the corresponding one, with the attribute of price range being preserved.

\end{document}